\documentclass{article}
\usepackage{spconf,amssymb,amsmath,graphicx}
\usepackage{caption}
\usepackage{multirow,makecell,hhline}
\usepackage{cite}
\usepackage{float}
\usepackage[fontsize=9pt]{fontsize}
\usepackage{bold-extra}
\usepackage[T1]{fontenc}
\usepackage{algorithm}
\usepackage{algorithmic}
\usepackage[subtle]{savetrees}

\usepackage{url}
\usepackage{hyperref}

\DeclareMathOperator*{\argmax}{arg\,max} 

\title{Back From The Future: Bidirectional CTC Decoding Using Future Information In Speech Recognition}
\name{Namkyu Jung, Geonmin Kim, Han-Gyu Kim}
\address{Naver Corporation, South Korea}

\begin{document}
%
\maketitle

\begin{abstract}
In this paper, we propose a simple but effective method to decode the output of Connectionist Temporal Classifier (CTC) model using a bidirectional neural language model. The bidirectional language model uses the future as well as the past information in order to predict the next output in the sequence. The proposed method based on bidirectional beam search takes advantage of the CTC greedy decoding output to represent the noisy future information. Experiments on the Librispeech dataset demonstrate the superiority of our proposed method compared to baselines using unidirectional decoding. In particular, the boost in accuracy is most apparent at the start of a sequence which is the most erroneous part for existing systems based on unidirectional decoding.
\end{abstract}


\section{Introduction}
In recent years, the performance of automatic speech recognition (ASR) systems have seen dramatic improvements due to the application of deep learning and the use of large-scale datasets \cite{dahl2011context, deng2013new, abdel2014convolutional, sak2014long, bahdanau2016end, amodei2016deep, zeghidour2018fully, chiu2018state, park2019specaugment}. In particular, there have been two strands of research pushing the state-of-the-art in this field, namely sequence-to-sequence models \cite{DBLP:journals/corr/ChanJLV15,dong2018speech} and Connectionist Temporal Classifier (CTC) based methods
 \cite{graves2006connectionist, graves2014towards}. This work focuses on the latter due to its simplicity and good performance. 
 
A key challenge in implementing such method lies in the process of decoding. Since CTC-based models do not have built-in methods for decoding, the beam search strategy is often used to achieve satisfactory accuracy. This requires an external language model (LM) to provide prior information on which sequence of words are the most probable. 


While both CTC and sequence-to-sequence models are able to process the whole sequence of speech data in a bidirectional manner, the traditional beam search method does not take advantage of the future information since the current prediction relies on the previous output. Although there have been attempts to take advantage of the whole sequence in the decoding process of CTC-based methods, its effectiveness has not been demonstrated in automatic speech recognition.

In this paper, we propose the bi-directional decoding method with noisy future context and training method robust to noisy context. 
As an ASR model, we employ the Connectionist Temporal Classification (CTC) model instead of autoregressive model because the latter is based on the \textit{label-synchronous} \cite{prabhavalkar2017comparison} decoding, resulting in difficulty to know which frame of speech is being decoded at each time step. However, \textit{frame-synchronous} \cite{hannun2014first, moritz2019streaming} decoding makes easier.

The majority of bi-directional LMs \cite{chen17, wang18, shin19} are designed for N-best LM rescoring with architectural advancement from LSTM-RNN to neural random field and transformer encoder. Although N-best LM rescoring can replace/be combined with our decoding method, we insist that rescoring generally underperforms shallow fusion. The former only finds the best solution among first-pass search results, while the latter can use LM score at every decoding time step of the beam search. 

\cite{zhang18} use the bidirectional shallow fusion decoding algorithm by using forward/backward uni-directional decoders on top of bi-directional encoder architecture. Similar to our work, they use the result of greedy decoding from the backward decoder as the future context. However, it is known that the greedy decoding of the autoregressive model suffers from exposure bias problem, which degrades the quality of generation and requires expensive supervision (e.g., scheduled sampling, REINFORCE) to fix it.

\vspace{-0.05in}
\section{CTC greedy and prefix beam search}
\label{sec:CTC prefix beam search}
\vspace{-0.05in}

Assume that sequence of speech representation $\mathbf{x}=\{x_1, \cdots, x_{T_x}\}$, corresponding sequence of labels $\mathbf{y}=\{y_1, \cdots, y_{T_y}\}$, and its alignment $\boldsymbol{\pi} = \{\pi_1, \cdots, \pi_{T_x} \}$ is given. The alignment sequence ($\boldsymbol{\pi}$) consists of a special CTC character called blank label as well as output labels in the set of label dictionary and is mapped to an unique result label sequence by removing all blanks and repeated labels.

For each time step $t$, the greedy decoding of the CTC model choose the most probable CTC output label using the probability $P(\pi_t|\mathbf{x})$, namely
\begin{equation} 
\label{eq:greedydecoding}
\boldsymbol{\pi}^* = \argmax_{\boldsymbol{\pi}}{\prod_{t}}P(\pi_t|\mathbf{x})
\end{equation}

Greedy decoding ends with having actual label sequence by being mapped to $\mathcal{B}$, $\mathbf{y}^* = B(\boldsymbol{\pi}^*)$. Greedy decoding is not able to get the most probable output sequence since it does assume that the probability of the sequence is independent for every frame. However it can be done very simply and quickly because the only thing have to do is find the most probable output label $\argmax_{\pi_t}P(\pi_t|\mathbf{x})$ for each time step $t$. 

Other than greedy decoding, we can efficiently calculate the probabilities of successive extensions of every each labelling prefix with prefix search decoding \cite{graves2006connectionist}. However, in full prefix search the maximum, number of prefixes expands grows exponentially as input sequence length gets longer. For that reason, prefix search decoding should be done with beam search decoding, which limits the size of extensions for each search step.

The CTC model is based on conditional independence assumption, therefore decoding often require an external language model to utilize dependency between outputs during decoding. Assume that we have an external language model which models the probability $P(y_t|y_{t-1}, \cdots y_1)$. CTC beam search can be done for a certain beam width $B$ by finding $B$ most probable candidates in each time step $t$. We should find the optimal CTC label sequence $\hat{\textbf{y}}=\mathcal{B}(\boldsymbol{\pi})$ such that
\begin{equation} \label{eq:ctcbeamsearch}
\argmax_{\boldsymbol{\pi}} \prod_t P_{\text{AM}}(\pi_t|\mathbf{x})P_{\text{LM}}(y_{\pi_t}|\mathcal{B}(\boldsymbol{\pi}_{1:t-1}))^{\alpha}
\end{equation}
where $P_{\text{AM}}$ is CTC probability, $P_{\text{LM}}$ is language model probability, $y_{\pi_t}$ is an output label of CTC label and $\alpha$ is hyperparameter controlling the effect of language model. 
For each time step $t$, get $B$ most probable beams based on the probability \eqref{eq:ctcbeamsearch} and proceed to next time step with $B$ beams.

\begin{figure}
    \centering
    \includegraphics[scale=0.15]{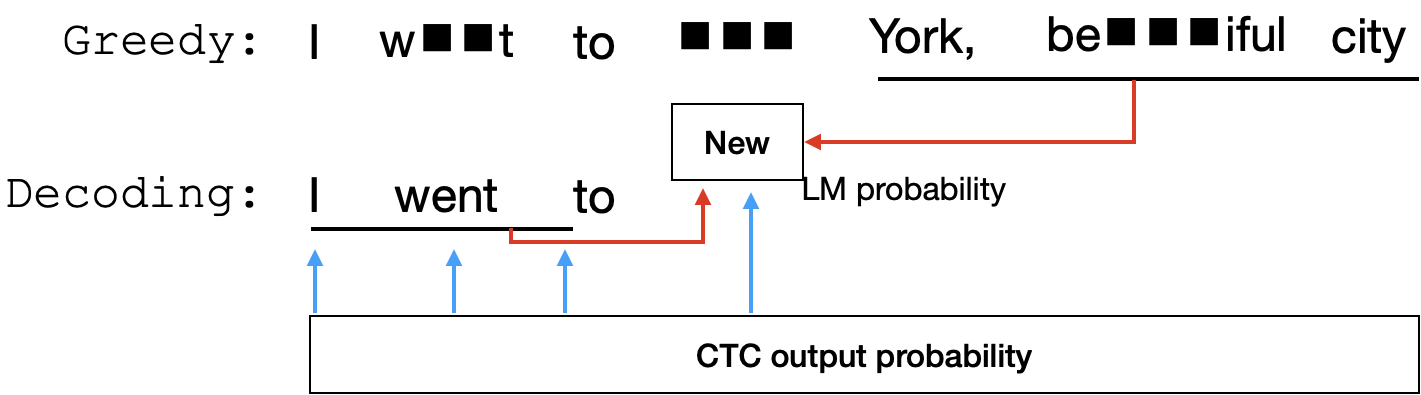}
    \caption{Bidirectional CTC Beam Search}
    \label{fig:bidecoding}
\end{figure}
\vspace{-0.1in}

\section{CTC Bidirectional Beam Search}
\label{sec:method}
\vspace{-0.05in}

\begin{figure*}[h!]
    \includegraphics[scale=0.3]{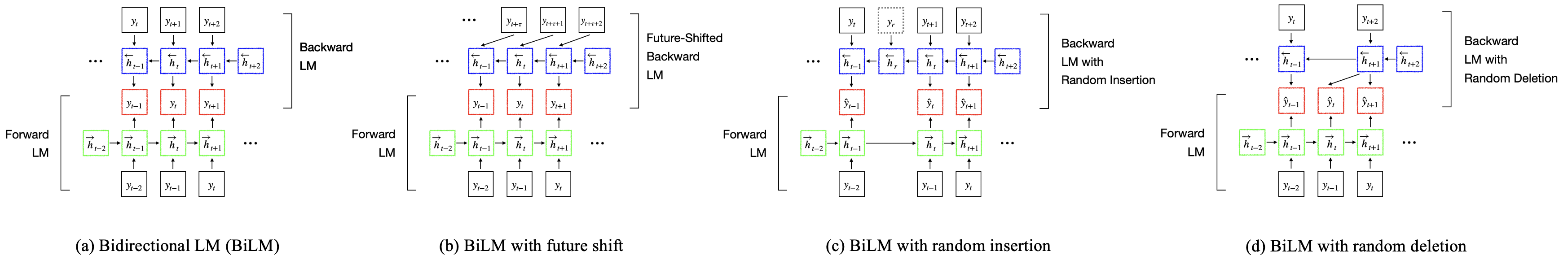}
    \caption{Bi-directional LM with future shift, random insertion/deletion noise}
    \label{fig:biLM_and_noise}
\end{figure*}

In this section, we propose bi-directional decoding method with noisy future context for CTC. In addition, training method making bi-directional language model robust to noisy future context is presented.

\subsection{Bidirectional Language Model}\label{subsec:lmfuture}
Assume that bi-directional context including future information is given. Then we can predict the next output token with both forward language model from the past information and backward language model from the future information with the following:
\begin{equation}\label{eq:bilm}
\begin{split}
P(y_t|y_1, \cdots, y_{t-1}, & y_{t+1}, \cdots, y_{T_y}) \\
&= \overleftrightarrow{f}_y(y_t|\overleftrightarrow{h}_{t}) \\
&= \overleftrightarrow{f}_y(y_t|\overrightarrow{h}_t + \overleftarrow{h}_t) \\
\end{split}
\end{equation}
where $\overrightarrow{h}_t = \overrightarrow{f}_h(y_{t-1} , \overrightarrow{h}_{t-1})$ represents the forward language model and $\overleftarrow{h}_t = \overleftarrow{f}_h(y_{t+1}, \overleftarrow{h}_{t+1})$ represents the backward language model respectively Fig. \ref{fig:biLM_and_noise}-(a). 


Bidirectional language model can be realized by Transformer as well as RNN with considering $h_t$ as memory in Transformer. We have used both LSTM and Transformer-XL to train bidirectional language model. It needs two input sequences, forward sequence for the past text and backward sequence for the future text. Putting into two input sequences, we get two hidden states each, $\overrightarrow{h}_t$ and $\overleftarrow{h}_t$ from bidirectional language model. $\overleftrightarrow{h}_{t}$ can be obtained by adding these two hidden states and it will be used to calculate final probability vector.

\subsection{CTC Decoding with Bidirectional Language Model}\label{subsec:bidecoding}
As described in Fig.\ref{fig:bidecoding}, while decoding through the time, greedy decoding result provides noisy future context for backward language model. In \eqref{eq:greedydecoding}, for greedy decoding result $\pi^*$, we may take $\mathcal{B}(\boldsymbol{\pi^*}_{t+1:})$ as the future sequence for a certain time step $t$. Therefore in order to decode CTC with bidirectional beam search, we find the optimal output sequence $\hat{\textbf{y}}=\mathcal{B}(\boldsymbol{\hat{\pi}})$ such that
\begin{equation} \label{eq:ctcbibeamsearch}
\begin{split}
\boldsymbol{\hat{\pi}} = \argmax_{\boldsymbol{\pi}} \prod_t & P_{\text{AM}}(\pi_t|\mathbf{x})\\
\times & P_{\text{BiLM}}(y_{\pi_t}|\mathcal{B}(\boldsymbol{\pi}_{1:t-1}), \mathcal{B}(\boldsymbol{\pi^*}_{t+1:}))^{\alpha}
\end{split}
\end{equation}
where $\alpha$ is the hyper-parameter controlling the effect of the language model to the final decoding and the second term of product can be evaluated as stated in $\eqref{eq:bilm}$.

With such bidirectional beam search, we may expect the decoding to get more valuable semantic information from the future sequence $\mathcal{B}(\boldsymbol{\pi^*}_{t+1:})$ helpful to predict next label output. Beam search with unidirectional language model can be relatively inaccurate at the start of the sentence because there are not much context for language model at the front part. Bidirectional language model having context from both front and back would be more grounded on support for the backward context from the future information.

\subsection{Future Exposure Bias}\label{subsec:exposurebias}
The well known \textit{exposure bias} problem implies that the model is exposed only on ground-truth context, while not exposed to predicted context. Both forward and backward language model (FWLM, BWLM) affect from this exposure bias problem due to its autoregressive nature. Especially the BWLM employ noisy context given from greedy decoding, where noisy context cannot be trivially obtainable from text corpus. To simulate noisy context given from CTC greedy decoding, we propose two data augmentation methods.

\subsubsection{Future Shift}\label{subsec:futureshift}
For backward language model, \textit{exposure bias} in the immediate future input can cause critical problem in predicting the next output. In order to mitigate this problem, we have to train the backward part in the bidirectional language model with a \textit{future shift} $\tau\in\mathbb{Z}$ so that there should be only effect of future sequence $\mathbf{y}_{t+1+\tau:}$ where $\mathbf{y}_{t+1:} = \mathcal{B}(\pi^*_{t+1:})$ on predicting $y_t$. If $\tau=0$, there is no shift in future and it means this backward may suffer from exposure bias problem. To achieve the expected result, backward language model should be trained and decoded with a future shift $\tau$. Fig.~\ref{fig:biLM_and_noise}-(b) shows how to make shift on future context.

\subsubsection{Random Noise In Future}
To further minimize the difference between training data and real data, we simulate noisy greedy decoding to be used in training backward language model. Greedy decoding could make three kinds of errors: insertion, deletion and substitution. Substitution error can be simulated by simply changing the input of a certain step, $y_t$ into some other random input $y'_t$. Simulating insertion and deletion is more complicated as both insertion and deletion break the alignment between the future sequence and input sequence. Fig.~\ref{fig:biLM_and_noise}-(c),(d) shows how to align future sequence with random noises when the bidirectional LM is trained.

\section{Experiments}
\label{sec:exps}
\vspace{-0.05in}

\subsection{Dataset}
The CTC is trained over the LibriSpeech dataset \cite{panayotov2015librispeech} which contains 960 hours of speech data of reading books. The models are tested in over two different datasets; \textit{dev-clean} and \textit{test-clean} which have 5.4 hours respectively. We used a character-level dictionary consisting of 48 tokens including special tokens such as \textit{unknown}, \textit{start of sentence}, \textit{end of sentence} and other special characters. Every model is trained on a
single NVIDIA V100 GPU.

The external bidirectional language model is trained on every train corpus consisting of about 280,000 sentences. Sentences shorter than 10 letters and consisting of almost non-alphabet characters have been removed. 

\subsection{Acoustic Model}

The 161-dimensional log spectrogram is used as the input of the CTC model. Two convolutional layers are used to compress the information of the log spectrogram on the temporal dimension. Four bidirectional LSTM layers with 2048 hidden nodes are used for acoustic information analysis. Then, one fully connected layer is used to compute the probability of each token from the output of the LSTM layers. The dimension of the final output vector of the CTC model is 49, which is one greater than the number of tokens because the CTC model should compute the probability of the \textit{blank} label.

\subsection{Language Model}
We used both LSTM and Transformer-XL in order to train bidirectional language model described in Section \ref{subsec:lmfuture}. We used 6 LSTM layers with implementation details following AWD-LSTM \cite{merity2017regularizing}. For Transformer-XL there are 6 Transformer decoder layers with memory-augmented hidden states for previous hidden states in auto-regressive decoding. Every hidden state has 1024-dimensional vector for each layer. 


There are two training parameters for bidirectional language model: future shift $\tau$ and random noise proportion $\epsilon$. We have experimented for $\tau=1, 2, 3$.
Another training parameter is $\epsilon$ controlling the random noise density in future sequence. We assume that the proportion of noise has to follow the actual error rate occurred in greedy decoding. Since the character error rate (CER) of CTC greedy decoding turned out around 5\%, we tried to set $\epsilon$ to be 0.05; 5 percents out of all characters in a sequence is set to be noise. Among these noises, we set 45\% to be insertion noise, and another 20\% to be deletion noise and the rest of them to be substitution noises, following the statistics of the greedy result Table \ref{table:1}. We are going to compare between the one with these modeling structure and the other without it. 

\begin{table}[h!]
\centering
\begin{tabular}{ |c|c|c| } 
    \hline
    type & dev-clean & test-clean \\ 
    \hline
    insertion & 43.2\% & 45.6\% \\ 
    deletion & 22.7\% & 21.9\% \\ 
    substitution & 34.1\% & 32.5\% \\ 
    \hline
\end{tabular}
\caption{Error types proportion in greedy decoding}
\label{table:1}
\end{table}


\subsection{Beam Search}
CTC prefix beam search finds the best $B$ beams for each time step $t$ with the score
\begin{equation}
\begin{split}
    score(\mathbf{y}) &= \sum_t \Big\{ \log P_{\text{CTC}}(y_t|X) \\
    &+ \alpha \log P_{\text{LM}}(y_t|y_{1:t-1}) \Big\} + \beta |\mathbf{y}|
\end{split}
\end{equation}
where $\alpha$ is the language model weight and $\beta$ is for the length reward for normalization with respect to the length due to the bias for shorter utterances.

For all experiments, we set $B=20$, $\alpha=1$, $\beta=2$ since we are looking for the effect of bidirectional decoding, not finding the optimal parameters. 

Detailed algorithm is described in Algorithm \ref{algorithm:beamsearch}. First, we get the greedy decoding result $\mathbf{y}_{1:T_y}^*$. Using it, we are able to obtain hidden states of backward language model for each $t$. These hidden states are cached and used for the bidirectional language model through time. For time $t$, we should get $t^*$ in line 10, the earliest time step after time $t$ having non-blank greedy output. With $t^*$ we can determine which part we are going to use among the backward hidden states. Depending on the future shift parameter $\tau$, $\overleftarrow{h}_{t^* + \tau}$ would be used as a backward hidden state. With forward hidden state $\overrightarrow{h}^b_{t}$, language model score $\boldsymbol{p}^{lm}$, $V$-dimensional vector having probability for each output label, can be calculated. By regular CTC prefix search algorithm (CTC\_PREFIX\_SEARCH) \cite{graves2006connectionist}, $\boldsymbol{p}^{total}$ is obtained, where the second argument of CTC\_PREFIX\_SEARCH is to be added to the non-blank probability in this function. With pruning beams over $\delta$, picking $B$ best beams based on the score $\boldsymbol{p}^{total}$ for the next time step. The inference of backward language model is done just once with greedy decoding result, there is not much difference in amount of total calculation.

\begin{algorithm}[t]                      
\caption{Bidirectional CTC beam search}
\label{algorithm:beamsearch}                      
\begin{algorithmic}[1]     
\REQUIRE $B$: \textrm{Beam Width}, $V$: \textrm{Vocab Size}, $\tau$: \textrm{future shift}, $\delta$: \textrm{Beam Threshold}, $\alpha$: \textrm{language model weight}, $\beta$: \textrm{length reward}
\STATE $\mathbf{y^*}_{1:T_y} \leftarrow \mathcal{B}(\pi^*_{1:T_x})$ (Greedy Decoding)
\STATE $\overleftarrow{h}_{T_y} \leftarrow \boldsymbol{0}$ \\
\FOR{$t \gets T_y-1$ \TO 0}
\STATE $\overleftarrow{h}_t \leftarrow \overleftarrow{f}_h(y^*_{t+1}, \overleftarrow{h}_{t+1})$
\ENDFOR

\STATE $\overrightarrow{h}_0 \leftarrow \boldsymbol{0}$
\STATE $\hat{B} \leftarrow [\textless s\textgreater ], \overrightarrow{h}^0_{0} \leftarrow \overrightarrow{h}_0$

\FOR{$t\gets1$ \TO $T_x$}
\STATE $\boldsymbol{p}^{\text{am}} \leftarrow \log P_{\text{CTC}}(y_t|\mathbf{X})$
\STATE \mbox{$t^* \leftarrow \min (\{s \in (t, T_y): \pi^*_s \neq \text{\textless blank\textgreater})\})$}
\FOR{$b\gets1$ \TO $|\hat{B}|$}
\STATE $\mathbf{y}^b \leftarrow \hat{B}[b]$
\STATE $\overrightarrow{h}^b_{t} \leftarrow \overrightarrow{f}_h(y_{t-1}, \overrightarrow{h}^b_{t-1})$
\STATE $\boldsymbol{p}^{\text{lm}} \leftarrow \log P_{\text{BiLM}} (y_t|\overrightarrow{h}^b_{t} + \overleftarrow{h}_{t^* + \tau})$
\STATE $\boldsymbol{p}^{total} \leftarrow \textrm{CTC\_PREFIX\_SEARCH}(\boldsymbol{p}^{am}, \alpha \boldsymbol{p}^{lm} + \beta |\mathbf{y}^b|)$
\FOR{$c\gets1$ \TO $V$}
\IF{$p^{total}_c > \delta$}
\STATE add $(c+\mathbf{y}^b)$ to $\hat{B}$
\ENDIF
\ENDFOR
\ENDFOR
\STATE $\hat{B}$ $\leftarrow$ top $B$ beams in $\hat{B}$
\STATE Reorder $\overrightarrow{h}^b_{t}$ according to $\hat{B}$
\ENDFOR

\end{algorithmic}
\end{algorithm}

\vspace{-0.1in}
\section{Results}
\label{sec:results}
\vspace{-0.05in}

\subsection{Perplexity}
In this section, we show the improvement of perplexity of the bidirectional language model compared to the unidirectional language model. We trained both LSTM and Transformer-XL varying the hyper-parameter $\tau$ from 1 to 3.

As we can see in Table \ref{table:2}, compared to the result of unidirectional language model, bidirectional LM has improved the model performance significantly for both LSTM and Transformer-XL based on the perplexity measure. This improvement is quite obvious since predicting with future information is surely more helpful than without it as well as future information is guaranteed to be perfect. Similarly, closer the information is, the more helpful it is because closer future is more related. Therefore smaller $\tau$ makes the results more accurate. However, like we explained in \ref{subsec:exposurebias}, smaller value of $\tau$ also increases the risk of inaccurate result from incorrect future information.

\begin{table}[h!]
\centering
\begin{tabular}{ |c|c|c|c|c| } 
    \hline
    model & \multicolumn{2}{|c|}{LSTM} & \multicolumn{2}{|c|}{T-XL} \\
    \hline
    dataset & dev & test & dev & test \\ 
    \hline
    uni &  5.811 & 5.745 & 5.539 & 5.505 \\ 
    $\tau$=1 & 2.711 & 2.676 & \textbf{2.451} & \textbf{2.245} \\ 
    $\tau$=2 & 4.327 & 4.281 & 3.409 & 3.214 \\ 
    $\tau$=3 & 5.351 & 5.171 & 4.564 & 4.452 \\ 
    \hline
\end{tabular}
\caption{Perplexity of Language Models}
\label{table:2}
\end{table}
\vspace{-0.1in}

\subsection{Decoding Results}

Table \ref{table:3} shows that for both datasets, bidirectional decoding is quite effective. However, we speculate that the direct future information gives side effects as we anticipated in advance, for observing the CER result of the case $\tau=1$. For LSTM $\tau=2$ seems to be optimal whereas $\tau=3$ for Transformer-XL. Furthermore, the best performance is obtained both for LSTM and Transformer-XL with adding random noise to greedy result during training with ratio of $\epsilon = 0.05$. We note that the purpose of our experiment is to prove the effectiveness of the proposed bidirectional decoding. Therefore, the mentioned $\tau$ and $\epsilon$ value may not be optimal. More experiments should be conducted in future to find out the generally optimal hyper-parameters.
\begin{table}[h!]
\centering
\begin{tabular}{ |c|c|c|c|c| } 
    \hline
    LM & \multicolumn{2}{|c|}{LSTM} & \multicolumn{2}{|c|}{T-XL} \\
    \hline
    dataset & dev & test & dev & test \\ 
    \hline
    greedy & 5.30 & 5.22 & 5.31 & 5.22 \\ 
    uni & 4.56 & 4.48 & 4.37 & 4.12 \\ 
    $\tau$=1 & 5.15 & 5.02 & 4.90 & 4.77\\ 
    $\tau$=2 & 4.40 & 4.23 & 4.29 & 4.12 \\ 
    $\tau$=3 & 4.51 & 4.31 & 4.22 & 4.05 \\ 
    \hline
    \makecell{$\tau$=2, $\epsilon$=0.05} & \textbf{4.38} & \textbf{4.21} & 4.24 & 4.07 \\ 
    \hline
    \makecell{$\tau$=3, $\epsilon$=0.05} & 4.46 & 4.28 & \textbf{4.17} & \textbf{4.02} \\ 
    \hline
\end{tabular}
\caption{Character Error Rate(CER) on LibriSpeech dev-clean and test-clean dataset for greedy search decoding, unidirectional beam search, and bidirectional beam search with $\tau=1, 2, 3$ and $\epsilon=0, 0.05$.}
\label{table:3}
\end{table}

\subsection{Relative Error Position}

\begin{figure}[h!]
    \includegraphics[scale=0.6]{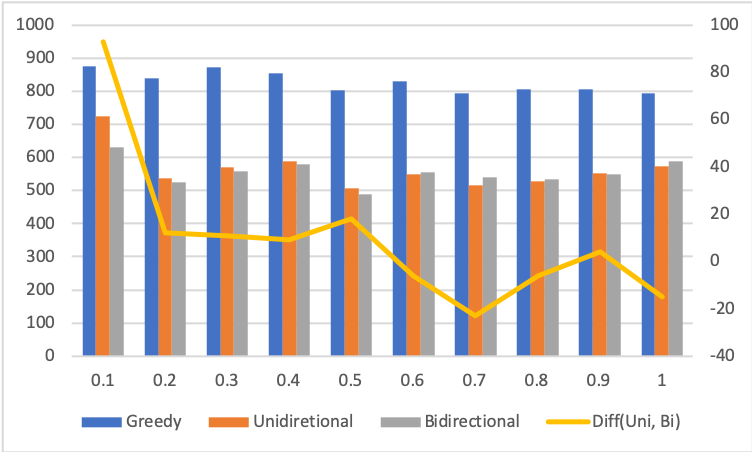}
    \caption{Histogram of relative positions where each error has occurred in \textit{test-clean} with Transformer-XL model and $\tau=3$: $x$-axis represents the relative position of every error part of the recognition results. 0 means the beginning of the speech and 1 means the end of the speech. $y$-axis shows the total count of errors for each bin. Yellow line implies the improvement from unidirectional to bidirectional.}
    \label{fig:chart}
\end{figure}

In order to further analyse the effectiveness of the proposed method, we visualized the statistics of the relative error position as a histogram in Fig.~\ref{fig:chart}. As shown in the figure, the unidirectional decoding shows obviously poorer performance at the beginning of the sentence, which coincides with the analysis in \ref{subsec:bidecoding}. On the contrary, bidirectional decoding shows relatively uniform distribution through all relative positions, implying that the proposed bidirectional decoding outperforms unidirectional one in overall error rates. nhancement gets the maximum at the front 10\% of the sequence, implying that the bidirectional decoding is especially good at correcting errors located in the front part.

In Table \ref{table:4}, there are a few examples that seems to be corrected by future information properly. For instance, the second sentence the greedy decoding result is \textit{tusday august eighteenth} and unidirectional decoding has no choice but to decode as it is. However for bidirectional decoding, it is able to see the future part \textit{august eighteenth} and successfully found out the right answer ``tuesday". Although the results are not flawless, but we can expect positive effect from the future like these examples.

\begin{table}[h!]
\centering
\begin{tabular}{ |m{11em}|m{11em}| } 
    \hline
    unidirectional & bidirectional \\
    \hline
    \small sweak squeak & \small \textbf{squeak} squeak \\ 
    \hline
    \small tusday august eighteenth & \small \textbf{tuesday} august eighteenth \\
    \hline
    \small i name nine others and said & \small i \textbf{named} nine others and said \\ 
    \hline
\end{tabular}
\caption{These examples show that bidirectional decoding uses future information  when necessary. In these cases, it is actually very hard to get the right text without seeing the future information as in unidirectional decoding.}
\label{table:4}
\end{table}

\vspace{-0.05in}
\section{Conclusions}
\label{sec:concl}
\vspace{-0.05in}

In this paper, we proposed new decoding method for CTC speech recognition model with language model trained bidrectionally. Bidirectional language model can be obtained by traditional unidirectional language model and backward language model for taking advantage of future information, for which greedy decoding result is used. With several experiments, we have demonstrated that the proposed method helps for decoding the front part of the speech compared to unidirectional decoding. Furthermore, we have alleviated future exposure bias problem by \textit{future shift} and \textit{random noise}. As a future work, we could apply bidirectional decoding to transducer-based model \cite{rao2017exploring, zeyer2021librispeech, liu2021improving} or label-synchronous attention-based decoder \cite{DBLP:journals/corr/ChanJLV15, gulati2020conformer}.

\vfill\clearpage

\bibliographystyle{IEEEbib}
\bibliography{shortstrings,refs}

\begin{thebibliography}{10}

\bibitem{dahl2011context}
George~E Dahl, Dong Yu, Li~Deng, and Alex Acero,
\newblock ``Context-dependent pre-trained deep neural networks for
  large-vocabulary speech recognition,''
\newblock {\em IEEE Transactions on audio, speech, and language processing},
  vol. 20, no. 1, pp. 30--42, 2011.

\bibitem{deng2013new}
Li~Deng, Geoffrey Hinton, and Brian Kingsbury,
\newblock ``New types of deep neural network learning for speech recognition
  and related applications: An overview,''
\newblock in {\em 2013 IEEE International Conference on Acoustics, Speech and
  Signal Processing}. IEEE, 2013, pp. 8599--8603.

\bibitem{abdel2014convolutional}
Ossama Abdel-Hamid, Abdel-rahman Mohamed, Hui Jiang, Li~Deng, Gerald Penn, and
  Dong Yu,
\newblock ``Convolutional neural networks for speech recognition,''
\newblock {\em IEEE/ACM Transactions on audio, speech, and language
  processing}, vol. 22, no. 10, pp. 1533--1545, 2014.

\bibitem{sak2014long}
Ha{\c{s}}im Sak, Andrew Senior, and Fran{\c{c}}oise Beaufays,
\newblock ``Long short-term memory based recurrent neural network architectures
  for large vocabulary speech recognition,''
\newblock {\em arXiv preprint arXiv:1402.1128}, 2014.

\bibitem{bahdanau2016end}
Dzmitry Bahdanau, Jan Chorowski, Dmitriy Serdyuk, Philemon Brakel, and Yoshua
  Bengio,
\newblock ``End-to-end attention-based large vocabulary speech recognition,''
\newblock in {\em 2016 IEEE international conference on acoustics, speech and
  signal processing (ICASSP)}. IEEE, 2016, pp. 4945--4949.

\bibitem{amodei2016deep}
Dario Amodei, Sundaram Ananthanarayanan, Rishita Anubhai, Jingliang Bai, Eric
  Battenberg, Carl Case, Jared Casper, Bryan Catanzaro, Qiang Cheng, Guoliang
  Chen, et~al.,
\newblock ``Deep speech 2: End-to-end speech recognition in english and
  mandarin,''
\newblock in {\em International conference on machine learning}, 2016, pp.
  173--182.

\bibitem{zeghidour2018fully}
Neil Zeghidour, Qiantong Xu, Vitaliy Liptchinsky, Nicolas Usunier, Gabriel
  Synnaeve, and Ronan Collobert,
\newblock ``Fully convolutional speech recognition,''
\newblock {\em arXiv preprint arXiv:1812.06864}, 2018.

\bibitem{chiu2018state}
Chung-Cheng Chiu, Tara~N Sainath, Yonghui Wu, Rohit Prabhavalkar, Patrick
  Nguyen, Zhifeng Chen, Anjuli Kannan, Ron~J Weiss, Kanishka Rao, Ekaterina
  Gonina, et~al.,
\newblock ``State-of-the-art speech recognition with sequence-to-sequence
  models,''
\newblock in {\em 2018 IEEE International Conference on Acoustics, Speech and
  Signal Processing (ICASSP)}. IEEE, 2018, pp. 4774--4778.

\bibitem{park2019specaugment}
Daniel~S Park, William Chan, Yu~Zhang, Chung-Cheng Chiu, Barret Zoph, Ekin~D
  Cubuk, and Quoc~V Le,
\newblock ``Specaugment: A simple data augmentation method for automatic speech
  recognition,''
\newblock {\em arXiv preprint arXiv:1904.08779}, 2019.

\bibitem{DBLP:journals/corr/ChanJLV15}
William Chan, Navdeep Jaitly, Quoc~V. Le, and Oriol Vinyals,
\newblock ``Listen, attend and spell,''
\newblock {\em CoRR}, vol. abs/1508.01211, 2015.

\bibitem{dong2018speech}
Linhao Dong, Shuang Xu, and Bo~Xu,
\newblock ``Speech-transformer: a no-recurrence sequence-to-sequence model for
  speech recognition,''
\newblock in {\em 2018 IEEE International Conference on Acoustics, Speech and
  Signal Processing (ICASSP)}. IEEE, 2018, pp. 5884--5888.

\bibitem{graves2006connectionist}
Alex Graves, Santiago Fern{\'a}ndez, Faustino Gomez, and J{\"u}rgen
  Schmidhuber,
\newblock ``Connectionist temporal classification: labelling unsegmented
  sequence data with recurrent neural networks,''
\newblock in {\em Proceedings of the 23rd international conference on Machine
  learning}, 2006, pp. 369--376.

\bibitem{graves2014towards}
Alex Graves and Navdeep Jaitly,
\newblock ``Towards end-to-end speech recognition with recurrent neural
  networks,''
\newblock in {\em International conference on machine learning}, 2014, pp.
  1764--1772.

\bibitem{prabhavalkar2017comparison}
Rohit Prabhavalkar, Kanishka Rao, Tara~N Sainath, Bo~Li, Leif Johnson, and
  Navdeep Jaitly,
\newblock ``A comparison of sequence-to-sequence models for speech
  recognition.,''
\newblock in {\em Interspeech}, 2017, pp. 939--943.

\bibitem{hannun2014first}
Awni~Y Hannun, Andrew~L Maas, Daniel Jurafsky, and Andrew~Y Ng,
\newblock ``First-pass large vocabulary continuous speech recognition using
  bi-directional recurrent dnns,''
\newblock {\em arXiv preprint arXiv:1408.2873}, 2014.

\bibitem{moritz2019streaming}
Niko Moritz, Takaaki Hori, and Jonathan Le~Roux,
\newblock ``Streaming end-to-end speech recognition with joint ctc-attention
  based models,''
\newblock in {\em Proc. IEEE Workshop on Automatic Speech Recognition and
  Understanding (ASRU)}, 2019.

\bibitem{chen17}
Chen Xie, Liu Xunying, Ragni Anton, Wang Yu, and Gales Mark,
\newblock ``Future word contexts in neural network language models,''
\newblock {\em IEEE Automatic Speech Recognition and Understanding}, 2017.

\bibitem{wang18}
Wang Bin and Ou~Zhijian,
\newblock ``Improved training of neural trans-dimensional random field language
  models with dynamic noise-contrastive estimation,''
\newblock {\em IEEE Spoken Language Technology}, 2018.

\bibitem{shin19}
Shin Joongbo, Lee Yoonhyung, and Jung Kyomin,
\newblock ``Effective sentence scoring method using bert for speech
  recognition,''
\newblock {\em ACML}, 2019.

\bibitem{zhang18}
Zhang Xiangwen, Su~Jinsong, Qin Yue, Liu Yang, Ji~Rongrong, and Wang Hongji,
\newblock ``Asynchronous bidirectional decoding for neural machine
  translation,''
\newblock {\em AAAI}, 2018.

\bibitem{panayotov2015librispeech}
Vassil Panayotov, Guoguo Chen, Daniel Povey, and Sanjeev Khudanpur,
\newblock ``Librispeech: an asr corpus based on public domain audio books,''
\newblock in {\em 2015 IEEE International Conference on Acoustics, Speech and
  Signal Processing (ICASSP)}. IEEE, 2015, pp. 5206--5210.

\bibitem{merity2017regularizing}
Stephen Merity, Nitish~Shirish Keskar, and Richard Socher,
\newblock ``Regularizing and optimizing lstm language models,''
\newblock {\em arXiv preprint arXiv:1708.02182}, 2017.

\bibitem{rao2017exploring}
Kanishka Rao, Ha{\c{s}}im Sak, and Rohit Prabhavalkar,
\newblock ``Exploring architectures, data and units for streaming end-to-end
  speech recognition with rnn-transducer,''
\newblock in {\em 2017 IEEE Automatic Speech Recognition and Understanding
  Workshop (ASRU)}. IEEE, 2017, pp. 193--199.

\bibitem{zeyer2021librispeech}
Albert Zeyer, Andr{\'e} Merboldt, Wilfried Michel, Ralf Schl{\"u}ter, and
  Hermann Ney,
\newblock ``Librispeech transducer model with internal language model prior
  correction,''
\newblock {\em arXiv preprint arXiv:2104.03006}, 2021.

\bibitem{liu2021improving}
Chunxi Liu, Frank Zhang, Duc Le, Suyoun Kim, Yatharth Saraf, and Geoffrey
  Zweig,
\newblock ``Improving rnn transducer based asr with auxiliary tasks,''
\newblock in {\em 2021 IEEE Spoken Language Technology Workshop (SLT)}. IEEE,
  2021, pp. 172--179.

\bibitem{gulati2020conformer}
Anmol Gulati, James Qin, Chung-Cheng Chiu, Niki Parmar, Yu~Zhang, Jiahui Yu,
  Wei Han, Shibo Wang, Zhengdong Zhang, Yonghui Wu, et~al.,
\newblock ``Conformer: Convolution-augmented transformer for speech
  recognition,''
\newblock {\em arXiv preprint arXiv:2005.08100}, 2020.

\end{thebibliography}

\end{document}